\documentclass[journal]{IEEEtranTIE}
\usepackage{graphicx}
\usepackage{cite}
\usepackage{picinpar}
\usepackage{amsmath}
\usepackage{url}
\usepackage{flushend}
\usepackage[latin1]{inputenc}
\usepackage{colortbl}
\usepackage{soul}
\usepackage{multirow}
\usepackage{pifont}
\usepackage{color}
\usepackage{alltt}
\usepackage[hidelinks]{hyperref}
\usepackage{enumerate}
\usepackage{siunitx}
\usepackage{breakurl}
\usepackage{epstopdf}
\usepackage{pbox}
\usepackage{bbm}
\usepackage{amssymb}
\usepackage{multirow,booktabs}
\usepackage{bm}
\usepackage{amsmath}

\begin{document}
\title{Precise Interception Flight Targets by Image-based Visual Servoing of Multicopter}

\author{
	\vskip 1em
	
	Hailong Yan, Kun Yang, Yixiao Cheng, Zihao Wang, and Dawei Li\\ 
	\thanks{
     Manuscript received September 24, 2024; revised March 7, 2025; accepted April 3, 2025. \textit{(Corresponding author: Dawei Li.)}  
		
     Hailong Yan is with the School of Computer Science, Northwestern Polytechnical University, Xi'an 710129, China. (e-mail: hlyan@mail.nwpu.edu.cn).
		
     Kun Yang, Yixiao Cheng, and Dawei Li are with the School of Automation Science and Electrical Engineering, the School of Future Aerospace Technology, and the Institute of Unmanned Systems, respectively, Beihang University, Beijing 100191, China. (e-mail: yangkun\_buaa@buaa.edu.cn; yixiao\_cheng@buaa.edu.cn; david@buaa.edu.cn).

     Zihao Wang is with the School of Cyber Security, University of Chinese Academy of Sciences, Beijing, 100049, China. (e-mail: wangzihao191@mails.ucas.edu.cn).}}

\maketitle
\begin{abstract}
Vision-based interception using multicopters equipped strapdown camera is challenging due to camera-motion coupling and evasive targets. This paper proposes a method integrating Image-Based Visual Servoing (IBVS) with proportional navigation guidance (PNG), reducing the multicopter's overload in the final interception phase. It combines smoother trajectories from the IBVS controller with high-frequency target 2D position estimation via a delayed Kalman filter (DKF) to minimize the impact of image processing delays on accuracy. In addition, a field-of-view (FOV) holding controller is designed for stability of the visual servo system. Experimental results show a circular error probability (CEP) of 0.089 m (72.8\% lower than the latest relevant IBVS work) in simulations and over 80\% interception success under wind conditions below 4 m/s in real world. These results demonstrate the system's potential for precise low-altitude interception of non-cooperative targets.
\end{abstract}


\markboth{IEEE TRANSACTIONS ON INDUSTRIAL ELECTRONICS}%
{}

\definecolor{limegreen}{rgb}{0.2, 0.8, 0.2}
\definecolor{forestgreen}{rgb}{0.13, 0.55, 0.13}
\definecolor{greenhtml}{rgb}{0.0, 0.5, 0.0}

Videos of the experiments: \href{https://youtu.be/sQdRCgnRp4g.}{https://youtu.be/sQdRCgnRp4g} and \href{https://youtu.be/vl2UTgEjeyk}{https://youtu.be/vl2UTgEjeyk}.

\section{Introduction}

\IEEEPARstart{T}{HE} presence of non-cooperative targets at low altitudes poses a significant threat to flight safety and hinders the development of low-altitude economies\cite{nature,paper01,paper02,paper03,paper04}. Commonly used countermeasures such as radio-frequency (RF) signal jamming\cite{paper05,paper06} and high-energy weapon shoot-downs\cite{paper07,paper08} are effective but have detrimental environmental impacts including signal pollution in the airspace and fires on the ground\cite{antisug,envent01}. On the other hand, capture methods\cite{PNG2024RAL,paper09,paper10} are environmentally friendly but less effective against moving targets. All these methods exhibit notable deficiencies when dealing with moving targets. The use of sensor-equipped unmanned aerial vehicles (UAVs) to intercept intruding targets has gained significant attention due to their rapid deployment, safety, and cost-effectiveness. Camera-based solutions, in particular, have shown great potential due to their low cost, lightweight, and high versatility\cite{paper12,paper11,paper13}.

\begin{figure}
    \centering
    \includegraphics[width=1\linewidth]{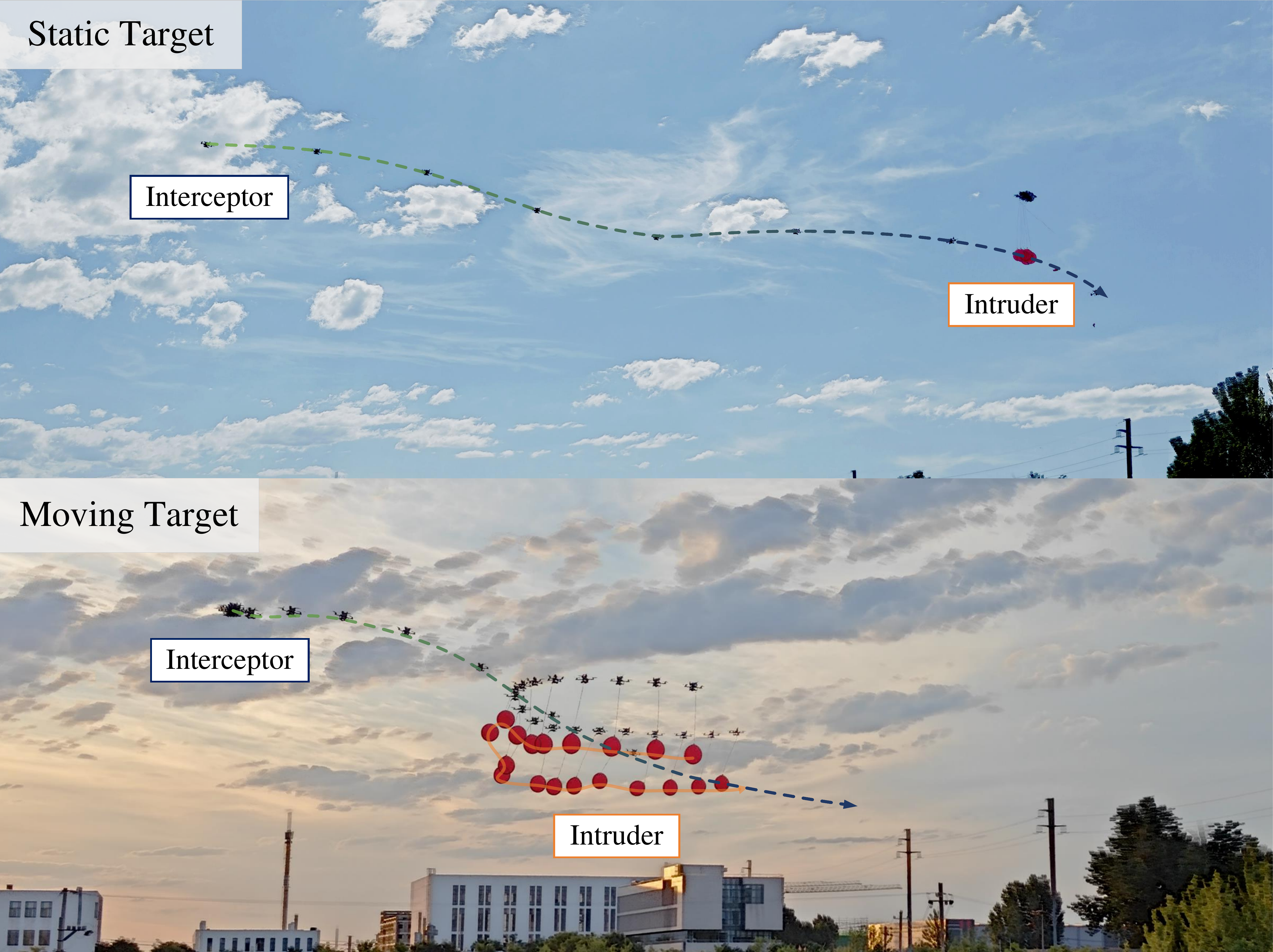}
    \caption{Intercepting flight target with multicopter.}
    \label{fig:1}
\end{figure}
In recent years, much research has focused on camera-equipped multicopters for intercepting non-cooperative targets. These methods are primarily divided into two categories: location estimation-based and IBVS-based. Common location estimation-based methods employ binocular cameras\cite{paper12,paper14}, the Kalman filter\cite{paper13,paper15}, and geometry-based approaches\cite{paper16} for target estimation. Subsequently, a trajectory for interception is planned. However, these methods are error-sensitive and limited by the sensor perception range and computational performance. Yet the IBVS methods have a simple structure that calculates the control quantity directly from the sensing module. Therefore it offers advantages such as insensitivity to modeling and calibration, fast response, and extended sensing capabilities. These advantages make it a highly competitive solution for interception tasks.

IBVS control of multicopter has been extensively studied in recent years and gradually there is some work for interception scenarios. Earlier methods support multicopter control such as stationary flight\cite{IBVS2008stable}, autonomous landing\cite{IBVS2012land} and translational rotation\cite{IBVS2014move} by extracting image features. Recent IBVS-based multicopter approaches tend to use only the target's bounding-box center point for airborne target tracking\cite{paper20,paper21} and interception\cite{paper18,paper22,paper19}. Some studies focus on innovations in target observation during tracking. A virtual-plane IBVS scheme \cite{paper21} decouples multicopter dynamics and uses an improved image error term to reduce the impact of multicopter rotation during 3D target chasing. A pseudo-linear Kalman filter \cite{paper20} and a 3D helical guidance law enhance observability for intruding multicopter tracking. In our previous work\cite{paper18,paper19}, four and six degree of freedom (DOF) IBVS controllers were designed for low and high speed target interception, respectively. Among them delayed Kalman filter (DKF) was proposed in \cite{paper19} to mitigate the delay in image processing. While some classic guidance laws, such as the pursuit guidance (PG) method \cite{paper22}, aim at lower off-target interception but are less efficient for moving targets. These methods can cause multicopter overload saturation or loss of the target due to image processing latency, inappropriate guidance strategies, and target maneuvers. No related work has focused on the adverse impact of overload on interception accuracy.

Multicopter IBVS faces significant challenges in precise interception of non-cooperative flight targets. During interception, the observable field of view (FOV) is susceptible to the multicopter's attitude, making it challenging to maintain the 2D visibility required for IBVS \cite{paper17}. Interception accuracy is further affected by target maneuvers, image processing delay, multicopter dynamics delay, and guidance strategy. While DKF \cite{paper19} addresses image processing delays, methods like PNG\cite{PNG,PNG2024JGCD,PNG2021TAES,PNG2024RAL} have been widely used to reduce overloads during the final interception phase. However, PNG is based on positional calculations, which may not be sufficient for non-cooperative target interception in real-world scenarios. Therefore, this paper designs a multicopter-friendly guidance law and controller capable of intercepting targets with low overload during the final interception phase, within the framework of existing IBVS sensors, using only the target angle information obtained from the monocular camera. By combining the above controller with our previously proposed DKF, we aim to achieve accurate interception of flight targets.

In this paper, an IBVS controller is designed for the precise interception of non-cooperative flight targets. This task bears a similarity to that of natural bird predators (see Fig. \ref{fig:1}). Specifically, the trajectory of a peregrine falcon hunting a mallard duck could be described by PNG \cite{paper23}, which mirrors the principles behind our approach. Moreover, a FOV holding controller mimics the bird's neck rotation for prey tracking. Thus, the proposed controller can continuously and stably track the target and achieve precise interception, similar to a peregrine falcon's hunting behavior. Overall, the contributions of this paper are summarized as follows:

\begin{enumerate}[1)]
     \item This work introduces a PNG-based IBVS controller that utilizes only monocular camera measurements for target information,  avoiding the assumptions of prior target knowledge \cite{PNG,PNG2021TAES,PNG2024JGCD} or the use of expensive and bulky LiDAR \cite{PNG2024RAL}.
    \item A controller for holding the field of view (FOV) is designed, incorporating PNG and multicopter dynamics to improve robustness in tracking targets under dynamic conditions.
    \item The proposed method outperforms the classical PG \cite{paper22} and state-of-the-art IBVS methods \cite{paper19,paper18} in interception accuracy for flight targets.
    \item The system's feasibility is validated through flight experiments, achieving a high interception success rate across diverse target motions and wind conditions.
\end{enumerate}

The rest of this paper is organized into five parts. Section \ref{sec:2} presents the basic modeling and algorithmic framework. Details of the IBVS controller design are provided in Section \ref{sec:3}. Simulation experiments are discussed in Section \ref{sec:4}, while real-world experiments are covered in Section \ref{sec:5}. Finally, Section \ref{sec:6} concludes with a summary of findings and directions for future research.

\section{MODELS AND FRAMEWORK}
\label{sec:2}
\begin{figure}[!ht]
    \centering
    \includegraphics[width=1\linewidth]{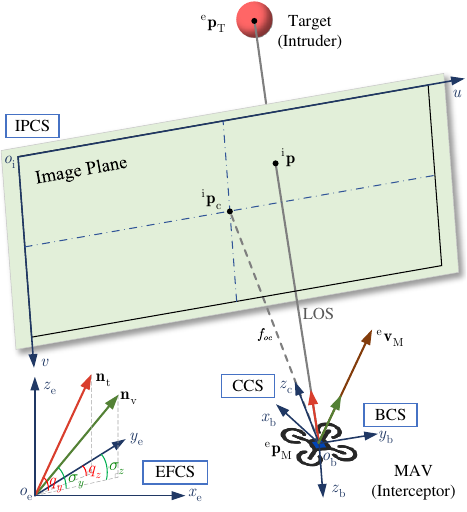}
    \caption{Coordinate system frames and description of the interception problem. In the earth coordinate system, vector $\mathbf{n}_{\mathrm{t}}$ (red vector) represents the LOS direction, while vector $\mathbf{n}_{\mathrm{v}}$ (green vector) represents the velocity direction. This information is obtained from sensors carried by the drone.}
    \label{fig:2}
\end{figure}
\subsection{Coordinate systems and relative motion models}
There are four coordinate systems used, as shown in Fig. \ref{fig:2}: the Earth-Fixed Coordinate System (EFCS): $o_e-x_{\mathrm{e}}y_ez_{\mathrm{e}}$, the Body Coordinate System (BCS): $o_b-x_by_bz_b$, Camera Coordinate System (CCS): $o_c-x_cy_cz_c$ and Image pixel coordinate system (IPCS): $o_i-uv$. 

In the relative motion model, both the target ($\mathrm{T}$) and the multicopter ($\mathrm{M}$) are assumed to be point masses. In the EFCS, the relative relationship can be expressed as:	
\begin{equation}
\begin{aligned}
^{\mathrm{e}}\mathbf{p}_{\mathrm{r}}&= {^{\mathrm{e}}\mathbf{p}_{\mathrm{T}}}-{^{\mathrm{e}}\mathbf{p}_{\mathrm{M}}}\\
^{\mathrm{e}}\mathbf{v}_{\mathrm{r}}&={^{\mathrm{e}}\mathbf{v}_{\mathrm{T}}}-{^{\mathrm{e}}\mathbf{v}_{\mathrm{M}}}\\
^{\mathrm{e}}\mathbf{a}_{\mathrm{r}}&={^{\mathrm{e}}\mathbf{a}_{\mathrm{T}}}-{^{\mathrm{e}}\mathbf{a}_{\mathrm{M}}}\\
&={^{\mathrm{e}}\mathbf{a}_{\mathrm{r}}^{\mathrm{t}}}+{^{\mathrm{e}}\mathbf{a}_{\mathrm{r}}^{\mathrm{n}}}
\end{aligned}
\end{equation}
where $^{\mathrm{e}}\mathbf{a}_{\mathrm{r}}^{\mathrm{t}}$ and $^{\mathrm{e}}\mathbf{a}_{\mathrm{r}}^{\mathrm{n }}$ represent the relative tangential and normal acceleration in the EFCS, respectively. The relative position $^{\mathrm{e}}\mathbf{p}_{\mathrm{r}}$ also denotes the line-of-sight (LOS) between the interceptor and the target.
\subsection{Multicopter Dynamics Model}

The multicopter used for interception is modeled as a rigid body with mass $m$. Its flight control rigid model can be summarized as \cite{paper24}:
\begin{equation}
\left. \left\{ \begin{aligned}
	^{\mathrm{e}}\dot{\mathbf{p}}_{\mathrm{M}}&={^{\mathrm{e}}\mathbf{v}_{\mathrm{M}}}\\
	^{\mathrm{e}}\dot{\mathbf{v}}_{\mathrm{M}}&=\mathbf{g}+\frac{1}{m} {^{\mathrm{e}}\mathbf{f}}\\
	\dot{\mathbf{R}}_{\mathrm{b}}^{\mathrm{e}}&=\mathbf{R}_{\mathrm{b}}^{\mathrm{e}}\left[ {^{\mathrm{b}}\bm{\omega}} \right] _{\times}\\
	\mathbf{J}\cdot {^{\mathrm{b}}\dot{\bm{\omega}}}&=-^{\mathrm{b}}\bm{\omega} \times \left( \mathbf{J}\cdot {^{\mathrm{b}}\bm{\omega}} \right) +\mathbf{G}_a+{^{\mathrm{b}}\mathbf{M}_{\mathrm{d}}}\\
\end{aligned} \right. \right. 
\end{equation}
where $^{\mathrm{e}}\mathbf{p}_{\mathrm{M}}$ and $^{\mathrm{e}}\mathbf{v}_{\mathrm{M}}$ is the multicopter's position and velocity in the EFCS, respectively. $\mathbf{g}$ denotes the acceleration due to gravity, typically represented as  $\left[\begin{array}{ccc} 0 & 0 & g \end{array}\right] ^{\mathrm{T}}$ with $g\approx$ 9.8 $\mathbf{m/s^2}$. $^{\mathrm{e}}\mathbf{f}\in\mathbf{R}^3$ is the controlled lift of the multicopter, directed opposite to the z-axis of the airframe. $\mathbf{R}_{\mathrm{b}}^{\mathrm{e}}\in \mathrm{SO} \left( 3 \right) $ represents the rotation from the BCS to the EFCS; $^{\mathrm{b}}\omega$ is the angular velocity in the BCS, and $\left[ ^{\mathrm{b}}\bm{\omega} \right] _{\times}$ is the cross-product matrix associated with $^{\mathrm{b}}\bm{\omega} $, where $\left[ \cdot \right] _{\times}$ denotes the matrix such that $\left[ \mathbf{x} \right] _{\times}\mathbf{y}=\mathbf{x}\times \mathbf{y}$ for any $\mathbf{x},\mathbf{y}\in \mathbf{R} ^3$. $\mathbf{J}$  is the moment of inertia of the multicopter, $\mathbf{G}_{\mathrm{a}}$ represents the gyroscopic moment, and $^{\mathrm{b}}\mathbf{M}_{\mathrm{d}}$ denotes the aerodynamic moment associated with propeller steering.

\subsection{Camera Image Model}
The multicopter is equipped with a strapdown monocular camera modeled using the pinhole model. This model projects target $^{\mathrm{c}}\mathbf{p}_{\mathrm{T}}=\left[ {^{\mathrm{c}}\mathrm{p}_{tx}}\,\,^{\mathrm{c}}\mathrm{p}_{ty}\,\,^{\mathrm{c}}\mathrm{p}_{tz} \right] ^{\mathrm{T}}$ from 3D space to the 2D camera plane: $\mathbbm{R} ^3\rightarrow \mathbbm{R} ^2$. The center point of the target in the IPCS is denoted as $^{\mathrm{i}}\mathbf{p}=\left[u,v \right]^{\mathrm{T}} $.The error in the IPCS, represented as $\mathbf{e}=\left[e_x,e_y\right]^{\mathrm{T}}$, is defined as:
\begin{equation}
\mathbf{e} \triangleq  {^{\mathrm{i}}\mathbf{p}}-{^{\mathrm{i}}\mathbf{p}_{\mathrm{c}}}
\end{equation}
where $^{\mathrm{i}}\mathbf{p}_{\mathrm{c}}=\left[ u_0,v_0 \right] ^{\mathrm{T}}$ is the geometric centroid of the image in the IPCS.  The relationship between the error and the change in camera motion can be described by the Jacobian matrix of the IBVS\cite{IBVS}: 
\begin{equation}
\label{eq:4}
\begin{aligned}
\dot{\mathbf{e}}&={\mathbf{L}_s}{^{\mathrm{c}}\tilde{\mathbf{v}}},\\
	&\mathbf{L}_s=\left[ \begin{matrix}
	-\frac{1}{^{\mathrm{c}}\mathrm{p}_{tz}}&		0&		\frac{\bar{e}_x}{^{\mathrm{c}}\mathrm{p}_{tz}}&		\bar{e}_x\bar{e}_y&		-\left( 1+\bar{e}_{x}^{2} \right)&		\bar{e}_y\\
	0&		-\frac{1}{^{\mathrm{c}}\mathrm{p}_{tz}}&		\frac{\bar{e}_y}{^{\mathrm{c}}\mathrm{p}_{tz}}&		1+\bar{e}_{y}^{2}&		-\bar{e}_x\bar{e}_y&		-\bar{e}_x\\
\end{matrix} \right]\\
\end{aligned}
\end{equation}
where $^{\mathrm{c}}\tilde{\mathbf{v}}=\left[ ^{\mathrm{c}}\mathbf{v}^{\mathrm{T}}\,\,^{\mathrm{c}}\mathbf{\omega }^{\mathrm{T}} \right] ^{\mathrm{T}}$ represents the instantaneous linear velocity angular and velocity of the camera. ${\bar{\mathbf{e}}}=\left[ \bar{e}_x\,\,\bar{e}_y \right] ^{\mathrm{T}}=\left[ e_x/f_{oc}\,\,e_y/f_{oc} \right] ^{\mathrm{T}}$ represents the
normalized image error of features in IPCS, and $f_{\mathrm{oc}}$ is the focal length of the camera.

The LOS vector direction $\mathbf{n}_{\mathbf{t}}=\left[ n_{tx}\,\,n_{ty}\,\,n_{tz} \right]^{\mathrm{T}} $, which connects the target and the multicopter in the EFCS, can then be characterized as:
\begin{equation}
   \begin{aligned}
	\mathbf{n}_{\mathbf{t}}&=\frac{^{\mathrm{e}}\mathbf{p}_{\mathrm{r}}}{\parallel {^{\mathrm{e}}\mathbf{p}_{\mathrm{r}}}\parallel}\\
	&\left. =\mathbf{R}_{\mathrm{b}}^{\mathrm{e}}\mathbf{R}_{\mathrm{c}}^{\mathrm{b}}\frac{[e_x\quad e_y\quad f_{\mathrm{oc}}]^{\mathrm{T}}}{\left\| \left[ \begin{matrix}
	e_x&		e_y&		f_{\mathrm{oc}}\\
\end{matrix} \right] \right. ^{\mathrm{T}}\parallel} \right.\\
\end{aligned} 
\end{equation}
where $\mathbf{R}_{\mathrm{c}}^{\mathrm{b}}$ is the rotation matrix from the CCS to the BCS. The LOS vector should not exceed the camera's FOV during interception to fulfill the 2D visibility of the IBVS.

\subsection{Framework overview}
\begin{figure}[!ht]
    \centering
    \includegraphics[width=1\linewidth]{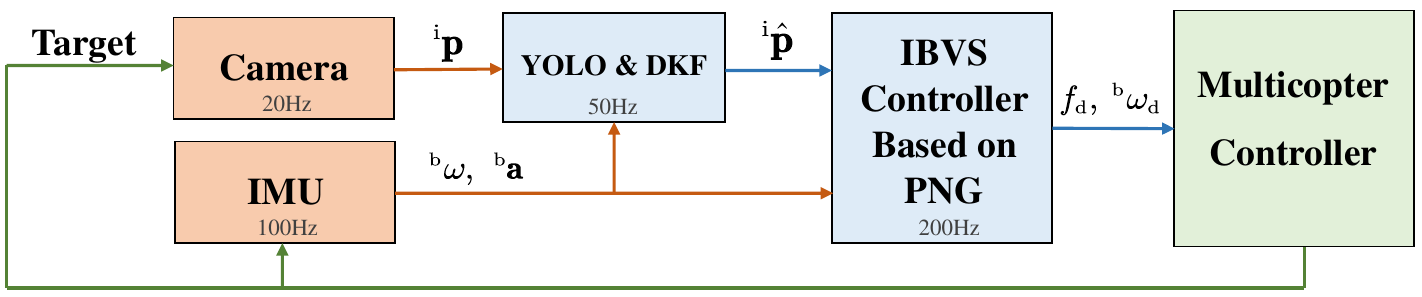}
    \caption{Framework of proposed IBVS controller based on PNG.}
    \label{fig:3}
\end{figure}

The system framework is shown in Fig. \ref{fig:3}, the blue blocks indicate the core components of the proposed algorithm. First, the image is processed using a target detection algorithm \cite{paper25}, and corrections are estimated using the DKF \cite{paper19}.  The IBVS controller algorithm then calculates the desired lift $f_{\mathrm{d}}$ and angular velocity $\omega _{\mathrm{d}}$ for the multicopter flight controller. Finally, multicopter flight controller labeled green, such as PX4, generates PWM signals for the motors to execute the target interception.

\section{CONTROLLER DESIGN AND ANALYSIS}
\label{sec:3}
\subsection{Guidance Algorithm Design}
The PNG calculates the rate of change of velocity proportional to the rate of change of the line of sight (LOS), causing the rate of change of the LOS to become progressively smaller and ultimately achieving interception:
\begin{equation}
\label{eq:5}
    \frac{\mathrm{d\sigma}}{\mathrm{dt}}=K\frac{\mathrm{dq}}{\mathrm{dt}}
\end{equation}
where $\sigma$ is the velocity angle, and $q$ is the LOS angle. $K$ is PNG constant. The trajectory before interception becomes progressively flatter and steadily reaches the target when the value exceeds a certain threshold\cite{PNG}. However, a value too large can make the controller aggressive and sensitive to noise. The value of $K$ is typically chosen between 2 and 6. 


The three-dimensional plane is divided into two two-dimensional planes, as shown in Fig. \ref{fig:2}. $q_y$ and $q_z$ are the LOS angles in the vertical and horizontal planes:
\begin{equation}\begin{aligned}&q_{y}=\arctan\left(\frac{n_{tz}}{\sqrt{n_{tx}^{2}+n_{ty}^{2}}}\right);q_{y}\in\left(-\frac{\pi}{2},\frac{\pi}{2}\right)\\&q_{z}=\arctan\left(\frac{n_{tx}}{n_{ty}}\right);q_{z}\in\left(-\frac{\pi}{2},\frac{\pi}{2}\right).\end{aligned}\end{equation}
And normalized velocity direction vector $\mathbf{n}_{\mathbf{v}}=\left[ n_{vx}\,\,n_{vy}\,\,n_{vz} \right]^{\mathrm{T}} $. Thus, $\sigma_y$ and $\sigma_z$ 
 are the velocity angles in the vertical and horizontal planes which are denoted as:
\begin{equation}\begin{aligned}&\sigma_{y}=\arctan\left(\frac{n_{vz}}{\sqrt{n_{vx}^{2}+n_{vy}^{2}}}\right);\sigma_{y}\in\left(-\frac{\pi}{2},\frac{\pi}{2}\right)\\&\sigma_{z}=\arctan\left(\frac{n_{vx}}{n_{vy}}\right);\sigma_{z}\in\left(-\frac{\pi}{2},\frac{\pi}{2}\right).\end{aligned}\end{equation}

Using the integral Eq. (\ref{eq:5}), desired velocity angle is calculated from the current moment $k$ and the previous moment $k-1$ as:
\begin{equation}
    \left\{ \begin{array}{c}
	\sigma _\mathrm{yd}=K_y(q_{y(k)}-q_{y(k-1)})+\sigma _{y(k-1)}\\
	\sigma _\mathrm{zd}=K_z(q_{z(k)}-q_{z(k-1)})+\sigma _{z(k-1)}\\
\end{array} \right. 
\end{equation}
where $K_y$ and $K_z$ are proportionality coefficient for the vertical and horizontal planes. Layman \textit{et al}. shows that a value of about 3 is appropriate when multirotor pursuit\cite{paperumsys}.

Then the desired velocity can be expressed as:
\begin{equation}
    \mathbf{v}_{\mathrm{d}}= v_{\mathrm{d}} \mathbf{n}_{\mathrm{vd}}
\end{equation}
where $\mathbf{n}_{\mathrm{vd}}=\left[ \cos \sigma _\mathrm{yd}\sin \sigma _\mathrm{zd}\,\,\cos \sigma _\mathrm{yd}\cos \sigma _\mathrm{zd}\,\,\sin \sigma _\mathrm{yd} \right] ^{\mathrm{T}}$. So the desired velocity can be get by designing the desired velocity magnitude.

\subsection{Field of View Holding Controller}
In this subsection, a FOV holding controller is designed to improve the robustness of IBVS during interception. Unlike some control strategies where the z-axis of the CCS is aligned with the target \cite{paper19} and the target converges to the center of the virtual image plane \cite{paper21}, a new strategy is proposed that aims to:
\begin{equation}
    \left\{ \begin{array}{c}
	e_x\rightarrow 0\\
	\varDelta e_y\leqslant \varepsilon\\
\end{array} \right. 
\end{equation}
Where $\varepsilon$ is a small constant and $\varDelta e_y$ is the difference between the largest and smallest $e_y$ during the interception. The primary control objective is to make the target converge to the center in the u-axis direction on the image plane. The secondary control objective is to minimize the range of motion in the v-axis direction.

The causes of target movement in the image plane during interception are analyzed. The main factors are the change of $^{\mathrm{c}}\mathbf{p}_{\mathrm{r}}$ due to relative motion and the change in attitude $\mathbf{R}_{\mathrm{b}}^{\mathrm{e}}$ of the multicopter. Due to the role of PNG, the LOS change $\varDelta q_{\mathrm{g}}$ is minimal. Therefore, the focus is on improving target tracking and minimizing the adverse effects of attitude changes by designing a suitable FOV holding controller.

Firstly, the relationship between camera motion and $e_x$ can represent from Eq. (\ref{eq:4}) as:
\begin{equation}
\label{eq:12}
\dot{e}_x=\left[ \frac{-1}{^{\mathrm{c}}\mathrm{p}_{tz}}\,\,0\,\,\frac{\bar{e}_x}{^{\mathrm{c}}\mathrm{p}_{tz}} \right] {^{\mathrm{c}}\mathbf{v}^{\mathrm{T}}}+\left[ \bar{e}_x\bar{e}_y\,\,-(1+{\bar{e}_x}^2) \,\,\bar{e}_y \right] {^{\mathrm{c}}\omega ^{\mathrm{T}}}.
\end{equation}
Due to the decoupling of the yaw angle and dynamics of the multicopter, a PD-based controller is designed for improving target tracking:
\begin{equation}
\label{eq:13}
^{\mathrm{b}}w_\psi=k_{\mathrm{p}}e_x+k_{\mathrm{d}}\dot{e}_x
\end{equation}
where $k_{\mathrm{p}}$ and $k_{\mathrm{d}}$ are small positive values. This controller improves the settling time and reduces the overshoot. Furthermore, it helps $e_x\rightarrow0$ and is insensitive to pixel noise.

Changes in roll angle at greater distances from the target have less impact on the target within the FOV. Therefore, primary attention is given to changes in the pitch angle. To minimize the adverse effects of attitude changes, an acceleration controller is designed as follows:
\begin{equation}
    {v}_{\mathrm{d}}  = v_{\mathrm{now}}  +k_a
\end{equation}
where $k_a$ is the velocity gain parameter, typically taken as 1-3. Thus, the dynamics-induced LOS error $\varDelta q_{\mathrm{d}}\leqslant \mathrm{arctan}(k_a/g)$. The change in the LOS angle of the target in the v direction of the IPCS can be expressed as:
\begin{equation}
    \varDelta q_\mathrm{y}=\varDelta q_{\mathrm{d}}+\varDelta q_{\mathrm{g}}.
\end{equation}

 Based on the geometric relationship, the variation of the target in the image coordinate system can be calculated as:
\begin{equation}
    \varDelta e_y\leqslant \frac{v_0\tan \left( \varDelta q_\mathrm
{y} \right)}{\tan \left( {\frac{1}{2}\alpha _\mathrm{vfov}} \right)}\leqslant \varepsilon 
\end{equation}
where the parameter $\alpha _\mathrm{vfov}$ is the angle of the camera's Vertical Field of View (VFOV).
\subsection{Attitude Loop Controller}
The desired acceleration is derived from the difference between the desired and current velocities:
\begin{equation}
    \mathbf{a}_{\mathrm{d}}=\frac{\mathbf{v}_{\mathrm{d}}-\mathbf{v}_{\mathrm{now}}}{\mathrm{dt}}
\end{equation}
where $\mathbf{v}_{\mathrm{now}}$ denotes the current velocity. Based on the multicopter's dynamics model and the equation $^{\mathrm{e}}\mathbf{f}_{\mathrm{d}}=f_{\mathrm{d}}\mathbf{n}_{\mathrm{fd}}$, the normalized direction of the lift force is:
\begin{equation}
\mathbf{n}_{\mathrm{fd}}=\frac{\mathbf{a}_{\mathrm{d}}-\mathbf{g}}{\parallel \mathbf{a}_{\mathrm{d}}-\mathbf{g}\parallel}.\textbf{}
\end{equation}

The attitude $\mathbf{R}_{\mathrm{d}}$, which incorporates only pitch and yaw, is designed to achieve the desired direction of lift. Here, the rotation matrix that includes only pitch and yaw angles is denoted as $\mathbf{R}_{\mathrm{title}}$. Thus, the desired attitude angle can be expressed as:
\begin{equation}
    \begin{aligned}
	\mathbf{R}_{\mathrm{d}}&=\mathbf{R}_{\mathrm{tilt}}\mathbf{R}_{\mathrm{b}}^{\mathrm{e}},\\
	\mathbf{R}_{\mathrm{tilt}}&=\mathbf{I}+[\mathbf{r}]_{\times}\sin \phi +\left[ \mathbf{r} \right] _{\times}^{2}\left( 1-\cos \phi \right)\\
\end{aligned}
\end{equation}
where $\mathbf{r}=\mathbf{n}_{\mathrm{f}}\times \mathbf{n}_{\mathrm{fd}}$, $\phi =\mathrm{arccos} \left( \mathbf{n}_{\mathrm{f}}^{\mathrm{ T}}\mathbf{n}_{\mathrm{fd}} \right) $. To achieve the desired pose as described above, a Lyapunov candidate function is formulated as:
\begin{equation}
    L_1=\mathrm{tr}\left( \mathbf{I}-\mathbf{R}_{\mathrm{d}}^{}\mathbf{R}_{\mathrm{b}}^{\mathrm{e}} \right) .
\end{equation}
Given that $\left\| \mathbf{I}-\mathbf{R}_{\mathrm{d}}^{\mathrm{T}}\mathbf{R}_{\mathrm{b}}^{\mathrm{e}} \right\| _{\mathrm{F}}=\sqrt{2\mathrm {tr}\left( \mathbf{I}-\mathbf{R}_{\mathrm{d}}^{\mathrm{T}}\mathbf{R}_{\mathrm{b}}^{\mathrm{e}} \right)}$, $L_1=\left\| \mathbf{I}-\mathbf{R}_{\mathrm{d}}^{\mathrm{T}}\mathbf{R}_{\mathrm{b}}^{\mathrm{e}} \right\| _{\mathrm{F}}^{2}/2\ge 0$, the derivative of the candidate function is:
\begin{equation}
    \begin{aligned}
\dot{L_1}&=-\mathrm{tr}\left(\mathbf{R}_{\mathrm{d}}^{\mathrm{T}}\mathbf{R}_{\mathrm{b}}^{\mathrm{e}}[^{\mathrm{b}}\bm{\omega}]_{\times} \right)
\\&=\mathrm{vex}\left(\mathbf{R}_{\mathrm{d}}^{\mathrm{T}} \mathbf{R}_{\mathrm{b}}^{\mathrm{e}}-\mathbf{R}_{\mathrm{b}}^{\mathrm{eT}}\mathbf{R}_{\mathrm{d}}\right)^{\mathrm{T}}
{^{\mathrm{b}}\bm{\omega}}.
\end{aligned}
\end{equation}
To make $\dot{L_1}\leqslant 0$ for stability, the pitch and roll angular velocity controllers are designed as:
\begin{equation}
    {^\mathrm{b}\bm{\omega}_1}=-\mathrm{vex}\left( \mathbf{R}_{\mathrm{d}}^{\mathrm{T}}\mathbf{R}_{\mathrm{b}}^{\mathrm{e}}-\mathbf{R}_{\mathrm{b}}^{\mathrm{eT}}\mathbf{R}_{\mathrm{d}} \right).
\end{equation}

In this context, the attitude control loop for the multi-rotor interceptor can be generalized:
\begin{equation}
    \begin{cases}
	^{\mathrm{b}}\bm{\omega} _{\mathrm{d}}=\mathrm{sat}\left( {^\mathrm{b}\bm{\omega}_1}+{{^\mathrm{b}\bm{\omega}_2},\omega _{\mathrm{m}}} \right)\\
	f_{\mathrm{d}}=\mathrm{min}\left( \mathrm{max}\left( \mathbf{n}_{\mathrm{f}}^{\mathrm{T}}\left( \mathbf{a}_{\mathrm{d}}-m\mathbf{g} \right) ,0 \right) ,f_{\mathrm{m}} \right)\\
\end{cases}
\end{equation}
where $^\mathrm{b}\bm{\omega}_1=\left[ ^\mathrm{b}w_{\psi}\,\,0\,\,0 \right] ^{\mathrm{T}}$. $\omega_{\mathrm{m}}$ and $f_{\mathrm{m}}$ denote the maximum angular velocity and maximum lift of the multicopter, respectively.  The saturation function of angular velocity $\mathrm{sat}\left( \cdot \right)$ is defined as:
\begin{equation*}
    \mathrm{sat}\left( \bm{\omega},w_{\mathrm{m}} \right) =\left\{ \begin{matrix}
	\bm{\omega},&\parallel \bm{\omega}\parallel \le w_{\mathrm{m}}\\
	\frac{w_m}{\parallel \bm{\omega}\parallel}\bm{\omega},&\parallel \bm{\omega}\parallel >w_{\mathrm{m}}\\
\end{matrix} \right. .
\end{equation*}
where it is considered to affect the convergence speed but not the system's stability\cite{paper19}.
\subsection{Stability Analysis of the Proposed Controller}
This subsection analyzes the closed-loop stability of the proposed image-based visual servo (IBVS) and proportional navigation guidance (PNG) fusion controller using Lyapunov's direct method.
In order to simplify the analysis while reflecting practical conditions, we make the following assumptions: 

1) Initial Target Visibility. The target is initially within the camera's field of view at a finite relative distance.

2) Target Maneuverability Limit. The target's maximum velocity and acceleration are bounded such that LOS angle changes are dominated by the interceptor's maneuvers.

Under these assumptions, we examine the tracking error dynamics and establish stability at the system level. We choose a Lyapunov candidate function as follows: 
\begin{equation}
\label{eq:24}
L_2 \;=\; \frac{1}{2}\Big(e_x^2 \;+\;^{\mathrm{e}}p_r^2\Big)\!
\end{equation}
where $^{\mathrm{e}}p_r=|^{\mathrm{e}}\mathbf{p}_\mathrm{r}|$ is the relative distance to the target in the inertial frame. The derivative of Eq. (\ref{eq:24}) is
\begin{equation}
\dot{L_2} \;=\; e_x\,\dot{e}_x \;+\; ^{\mathrm{e}}p_r\,{}^{\mathrm{e}}\dot{p}_r\,.
\end{equation}
By construction, $L_2 \ge 0$ for all defined error states, and $L_2 = 0$ if and only if $e_x = {}^{\mathrm{e}}p_r = 0$. According to Lyapunov's second method, in order to make the system stable, $\dot{L_2}$ is expected to be negative definite.

For the image-plane error component, we approximate Eq. (\ref{eq:12}) as \cite{paper18}
\begin{equation}
\label{eq:26}
\dot{e}_x= \frac{\bar{e}_x}{^{\mathrm{c}}\mathrm{p}_{tz}} {^{\mathrm{c}}v_z}-(1+{\bar{e}_x}^2){^{\mathrm{c}}\omega_y}.
\end{equation}
 Considering ${^{\mathrm{c}}\omega_y}\approx{^{\mathrm{b}}w_\psi}$, substituting Eq. (\ref{eq:13}) into Eq. (\ref{eq:26}) yields a relationship between $e_x$ and $\dot{e}_x$ as
\begin{equation}
\dot{e}_x=-\lambda {e}_x
\end{equation}
where
\begin{equation*}
\lambda=\frac{-\frac{{^\mathrm{c}v_z}}{{f_{oc}}^\mathrm{c}p_{tz}}+k_p\left( 1+\bar{e}_{x}^{2} \right)}{1+k_d\left( 1+\bar{e}_{x}^{2} \right)}. 
\end{equation*}
In order to make $\lambda>0$, we have make $k_d>0$ and $k_p>\frac{^\mathrm{c}v_z}{f_{oc}{^\mathrm{c}p_{tz}}\left( 1+\bar{e}_{x}^{2} \right)}$. That is, $e_x$ and $\dot{e}_x$ always have opposite signs. Therefore, the first term in $\dot{L_2}$, $e_x\,\dot{e}_x$, is strictly negative.

Next, we examine the relative distance term. Neglecting the accelerations of both the interceptor and the target for analysis tractability, the LOS angular rate can be expressed by the relative motion geometry as 
\begin{equation}
\label{eq:28}
    {}^{\mathrm{e}}p_r\,\dot{q} \;=\; v_m \sin\eta_m \;-\; v_t \sin\eta_t\
\end{equation}
where $\eta_m = q - \sigma_m$ and $\eta_t = q - \sigma_t$ are the interceptor's and target's velocity leading angles, respectively. Differentiating the above relation and applying the PNG law $\dot{\sigma}_m = K\,\dot{q}$ for the interceptor yields 
\begin{equation}
    {}^{\mathrm{e}}p_r\,\ddot{q} \;=\; -\Big(K\,v_m \cos\eta_m \;+\; 2\,{}^{\mathrm{e}}\dot{p}_r\Big)\,\dot{q}\,.
\end{equation}
If the navigation constant $K$ is chosen to satisfy 
\[ 
K \;>\; \frac{2\,\big|{}^{\mathrm{e}}\dot{p}_r\big|}{\,v_m \cos\eta_m\,}\,, 
\] 
then the coefficient $\big(K\,v_m \cos\eta_m + 2\,{}^{\mathrm{e}}\dot{p}_r\big)$ becomes positive. In this case, $\ddot{q}$ is always of opposite sign to $\dot{q}$, which means the LOS rate $\dot{q}$ will monotonically decrease in magnitude and converge to zero. Therefore, $v_m \sin\eta_m-v_t \sin\eta_t$ of Eq. (\ref{eq:28}) converge to zero. Additionally, base on target maneuverability limit that $v_t<v_m$ and initial target visibility that $\cos\eta_m>0$, we can easily proved that \cite{TIEPNG}:
\begin{equation}
 {}^{\mathrm{e}}\dot{p}_r \;=\; v_t \cos\eta_t \;-\; v_m \cos\eta_m\,<0. 
\end{equation}
Since ${}^{\mathrm{e}}p_r > 0$ by definition and ${}^{\mathrm{e}}\dot{p}_r < 0$, the second term $^{\mathrm{e}}p_r\,{}^{\mathrm{e}}\dot{p}_r$ in $\dot{L_2}$ is also negative during the engagement.

In summary, since $L_2$ is positive definite and $\dot{L_2} < 0$ for all nonzero errors, Lyapunov's second method ensures the closed-loop error system asymptotically stable, guaranteeing the interceptor's stable convergence to the target. However, the stability conclusions in this paper rely on the assumptions of limited target maneuverability and superior interceptor speed, with performance degradation observed under high-speed target escape conditions, as evidenced by experiments.

\section{SITL EXPERIMENTS}
\label{sec:4}
\subsection{Platform Introduction and Experimental Design}
In this section, the proposed algorithm is validated and compared using SITL simulation experiments with RflySim \cite{paper26}.These simulations replicate the architecture and algorithms of real flight experiments, ensuring high fidelity and ease of portability. Initially, a multi-directional static target interception experiment is designed to verify the accuracy of the proposed algorithm. Next, moving target interception experiments are performed and compared with other algorithms. These experiments include targets with three common motion models and initial relative positions in three different directions.
\begin{figure*}[!ht]
    \centering
    \includegraphics[width=0.65\linewidth]{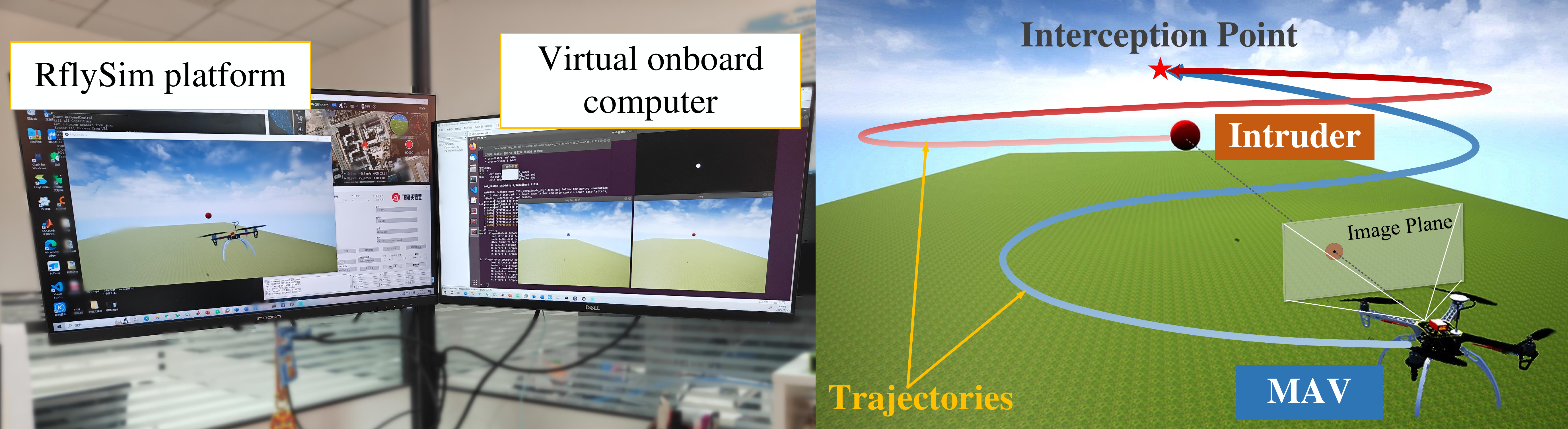}
    \caption{SITL Platform and Simulation Scenario.}
    \label{fig:4}
\end{figure*}

\subsection{Static Target Intercept Simulation Experiments}
\begin{figure}
    \centering
    \includegraphics[width=0.9\linewidth]{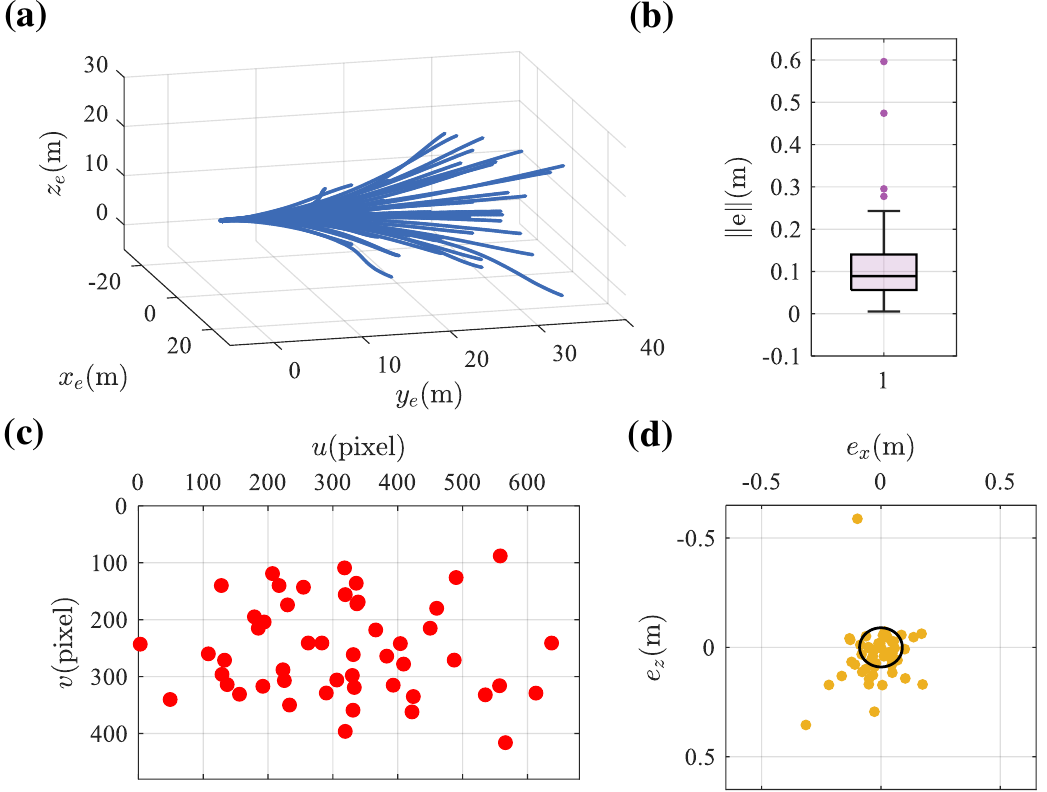}
    \caption{Experiment results of static target interception simulation: (a) multicopter trajectories, (b) boxplots of interception error, (c) initial 
 tagert positions in image plane, and (d) interception error distribution.}
    \label{fig:5}
\end{figure}

Fifty static target interception experiments are performed with targets at various locations. The multicopter starts at the position (0,0,10) meters in the EFCS, while the target positions, as shown in Fig. \ref{fig:5}(a), are randomly generated between 15 and 35 meters from the multicopter. The multicopter's initial velocity is set to move directly forward, and $k_a=2$, $K_y=K_z=3$, $k_p=0.03$, $k_d=0.01$ are set.

The recorded data from the proposed controller during static target interception are shown in Fig. \ref{fig:5}(a)--(d). The distributions of 50 targets in 3D space and the image plane are shown in (a) and (c), respectively. Interception accuracy is quantified by the Circular Error Probability (CEP), which indicates the radius within which the interceptor has a 50\% probability of hitting the target \cite{paper27}. The CEP is $\mathrm{CEP}_{\mathrm{Proposed}} = 0.089$ m, as shown in (b) and (d). This represents a substantial improvement over previous works \cite{paper18} and \cite{paper19}, reporting CEP values of $\mathrm{CEP}_{\left[ 26 \right]} = 0.457$ m and $\mathrm{CEP}_{\left[ 28 \right]} = 0.332$ m, respectively. The proposed method achieves reductions of 80.5\% and 73.2\%, demonstrating a significant improvement in accuracy. Notably, this result marks a transition from decimeter-level performance to centimeter-level accuracy in this classic test scenario.

 \subsection{Moving Target Intercept 
 Simulation Experiments}
\begin{figure}[!ht]
    \centering
    \includegraphics[width=1\linewidth]{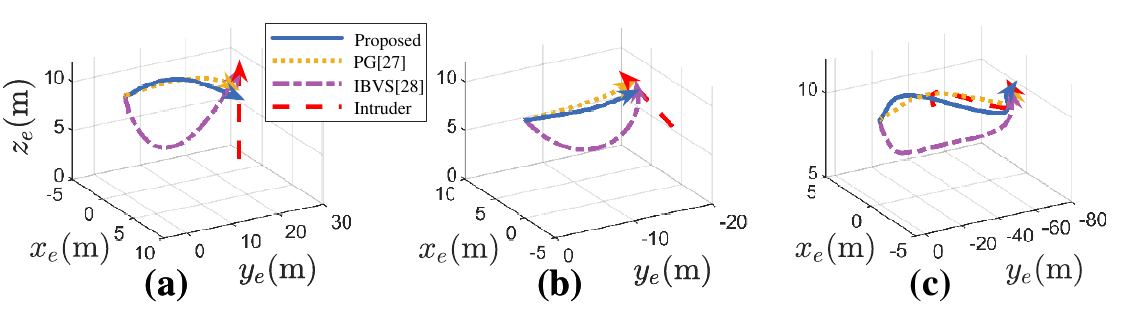}
    \caption{Trajectory of (a) CV, (b) CA, and (c) SM maneuvering model during moving target interception simulation.}
    \label{fig:6}
\end{figure}

This experiment evaluates the interception accuracy of the proposed algorithm across various target maneuvering scenarios. Three maneuvering models were tested: constant velocity (CV), constant acceleration (CA), and sinusoidal maneuver (SM), each modeling different types of evasive actions typically encountered in low-altitude target interception tasks. Parameters were consistent with those used for intercepting stationary targets. The initial relative positions included three interception scenarios: top-to-bottom, flanking, and tailgate, as shown in Fig. \ref{fig:6}. These scenarios test the algorithm's performance in diverse dynamic environments, challenging the controller's robustness in both direct pursuit and evasive maneuvers. Initially, the interceptor multicopter hovers at (0, 0, 10) m. The target follows three different maneuvering models starting from (0, 25, 1) m, (-8, 15, 8) m, and (0, 30, 10) m, respectively. The models are expressed as:
\begin{equation*}
\begin{cases}
	\mathrm{CV}:{^\mathrm{e}\mathbf{v}_{\mathrm{T}}}=\left( 0,0,1 \right)\\
	\mathrm{CA}:{^\mathrm{e}\mathbf{a}_{\mathrm{T}}}=\left( 0.8,0,0.2 \right)\\
	\mathrm{SM}:{^\mathrm{e}\mathbf{p}_{\mathrm{T}}}=\left( 2\sin\mathrm{(}2\pi /14t),3t,0 \right).\\
\end{cases}
\end{equation*}
\begin{figure}[!ht]
    \centering
    \includegraphics[width=1\linewidth]{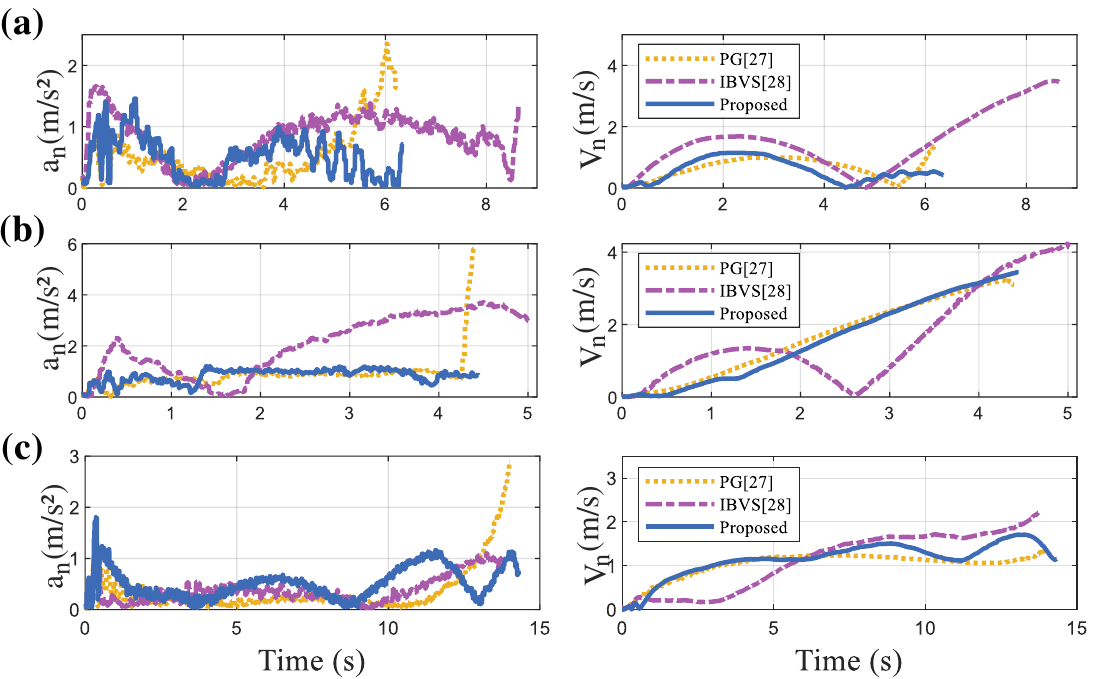}
    \caption{Normal velocity and normal acceleration of (a) CV, (b) CA, and (c) SM maneuvering model during moving target interception simulation.}
    \label{fig:7}
\end{figure}

\begin{figure}[!ht]
    \centering
    \includegraphics[width=0.92\linewidth]{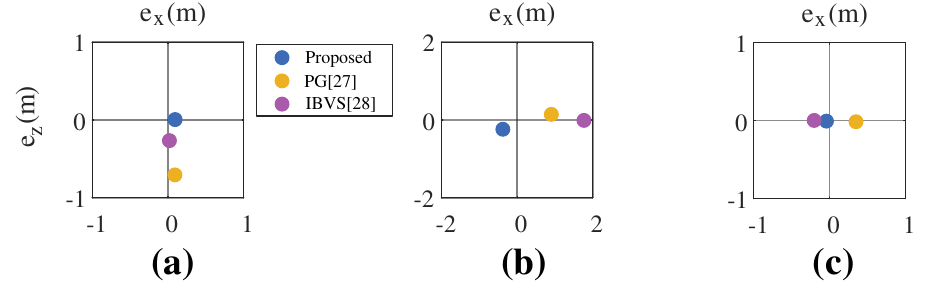}
    \caption{Errors of (a) CV, (b) CA, and (c) SM maneuvering model in moving target interception simulation.}
    \label{fig:8}
\end{figure}

\begin{figure}[!ht]
    \centering
    \includegraphics[width=0.85\linewidth]{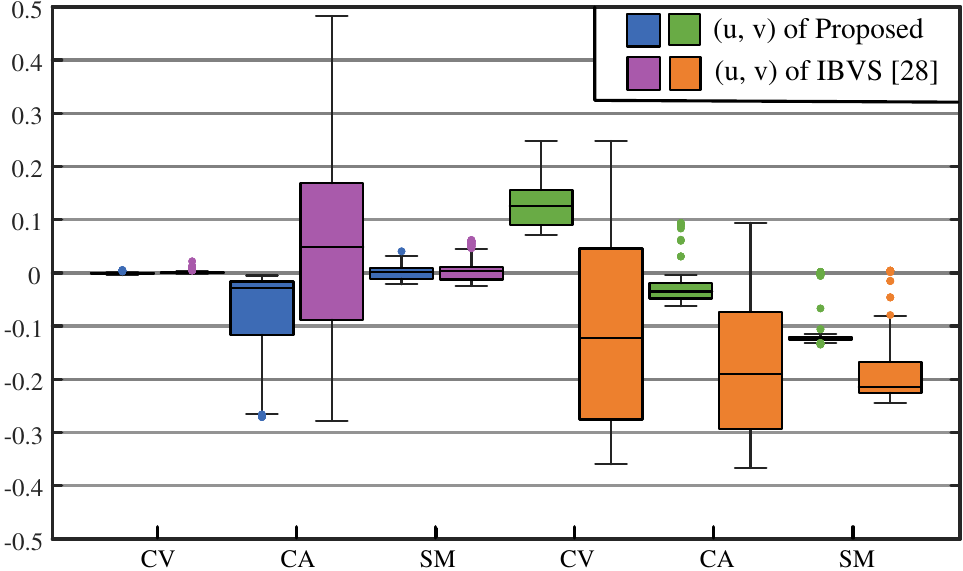}
    \caption{Normalized image plane error comparison.}
    \label{fig:9}
\end{figure}

As shown in Fig. \ref{fig:6}, the proposed algorithm achieves smoother trajectories when intercepting moving targets. The lower normal acceleration and velocity in the final phase of interception, shown in Fig. \ref{fig:7}, demonstrate the effectiveness of PNG in the proposed IBVS controller. Interception errors are illustrated in Fig. \ref{fig:8}. Although the trajectory smoothness and normal velocity of PG are lower than those of the proposed method in some scenarios, PG experiences a surge in normal acceleration towards the end, leading to reduced interception accuracy, as shown in Fig. \ref{fig:7} and Fig. \ref{fig:8}. Table \ref{tab:1} confirms that the proposed algorithm's interception accuracy outperforms the other two algorithms in intercepting moving targets.
\begin{table}[!ht]
	\centering
	\caption {Interception error of experiments.}
	\label{tab:1}
	\begin{tabular}[c]{*{4}{c}}
		\toprule
		Experimental indicators& PG\cite{paper22} &IBVS \cite{paper19}&Proposed \\
            \midrule
		{Errors of CV model (m)} & 0.71 & 0.27 &\textbf{0.09} \\
		\midrule
	{Errors of CA model (m)} & 0.92 & 1.76 &\textbf{0.45} \\
  
  		\midrule
		{Errors of SM model (m)} & 0.31 &0.20&\textbf{0.04} \\
		\bottomrule
	\end{tabular}
\end{table}
Additionally, the effectiveness of the FOV holding controller is evaluated by comparing it with another algorithm. The median, quartiles, boundaries, and outliers of the target in the image plane are shown in Fig. \ref{fig:9}. The proposed method exhibits a smaller error variation range than the compared algorithm \cite{paper19}, indicating superior field of view holding capability. This demonstrates the proposed algorithm's greater potential to handle more complex target maneuver scenarios.

\section{REAL FLIGHT EXPERIMENTS}
\label{sec:5}

\subsection{Introduction to Flight Experiments}
Extensive SITL experiments demonstrate the interception accuracy and effectiveness of the proposed algorithm. In this section, a series of experiments will be conducted to verify the effectiveness and robustness of the proposed algorithm in different real scenarios.

In all experiments, the parameter settings were kept consistent with the simulation: $k_a=2$, $K_y=K_z=3$, $k_p=0.03$, $k_d=0.01$, and the red balloon was used instead of the target.
The indoor flight experiments under different wind speeds were designed by first considering the potential effects of outdoor wind conditions on both the target balloon and the vehicle. Following this, the experimental scenarios were set outdoors. Static target interception, moving target interception, and comparison experiments were conducted.
\subsection{Indoor Experiments with Different Wind Conditions}
\begin{figure}[!ht]
    \centering
    \includegraphics[width=0.90\linewidth]{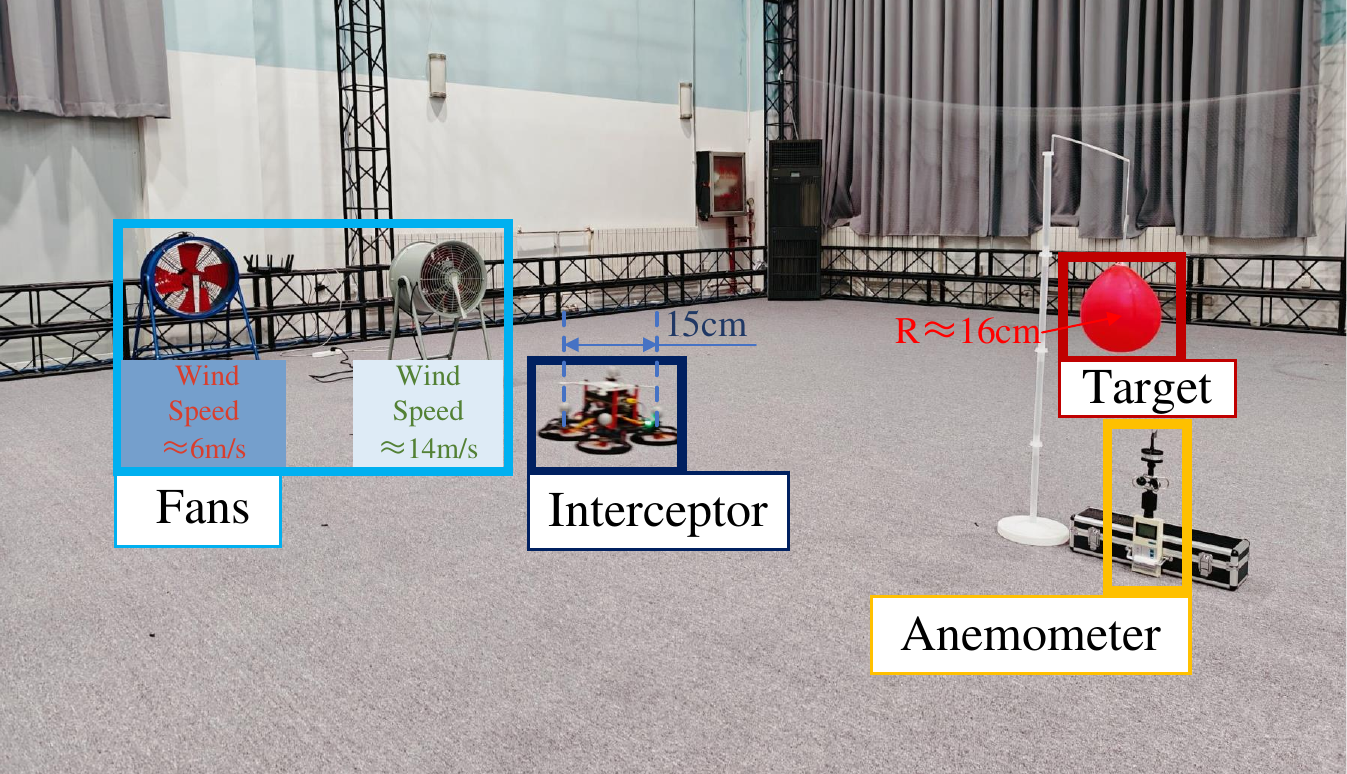}
    \caption{Indoor interception experiment scenarios under different wind conditions.}
    \label{fig:17}
\end{figure}

The indoor experiments were conducted using the experimental platform shown in Fig. \ref{fig:17}. The multicopter is equipped with a Jetson CSI monocular camera ($120^\circ$ FOV) for capturing images, a NVIDIA Jetson Xavier NX for processing sensor data, and a Pixhawk 6C Mini for flight control. The fans can provide winds of 2 to 6 m/s from different distances, and the target balloon is attached to the base by a soft rope to swing with the wind. The target balloon had a diameter of about 16 cm, and the multi-rotor drone had a wheelbase of 15 cm, making it a very challenging series of tests. A total of 20 interception experiments were conducted at different wind speeds. The video of the experiments can be found at \href{https://youtu.be/vl2UTgEjeyk}{https://youtu.be/vl2UTgEjeyk} or \href{https://b23.tv/cx5egf4}{https://b23.tv/cx5egf4}.

To ensure experimental consistency, the same balloon was reused across all tests, with the Optitrack motion capture system providing accurate initial position data for the interceptor. The multicopter began at the position (0, 6, 1.5) m and intercepted the balloon located at (0.5, 2.3, 1) m. Wind speed at the balloon's location was measured using an anemometer when the balloon was stationary. A series of five consecutive experiments were conducted at wind speeds of 0 m/s, 2 m/s, 4 m/s, and 6 m/s to evaluate the robustness of the controller and the interception success rate.

\begin{figure}[!ht]
    \centering
    \includegraphics[width=1\linewidth]{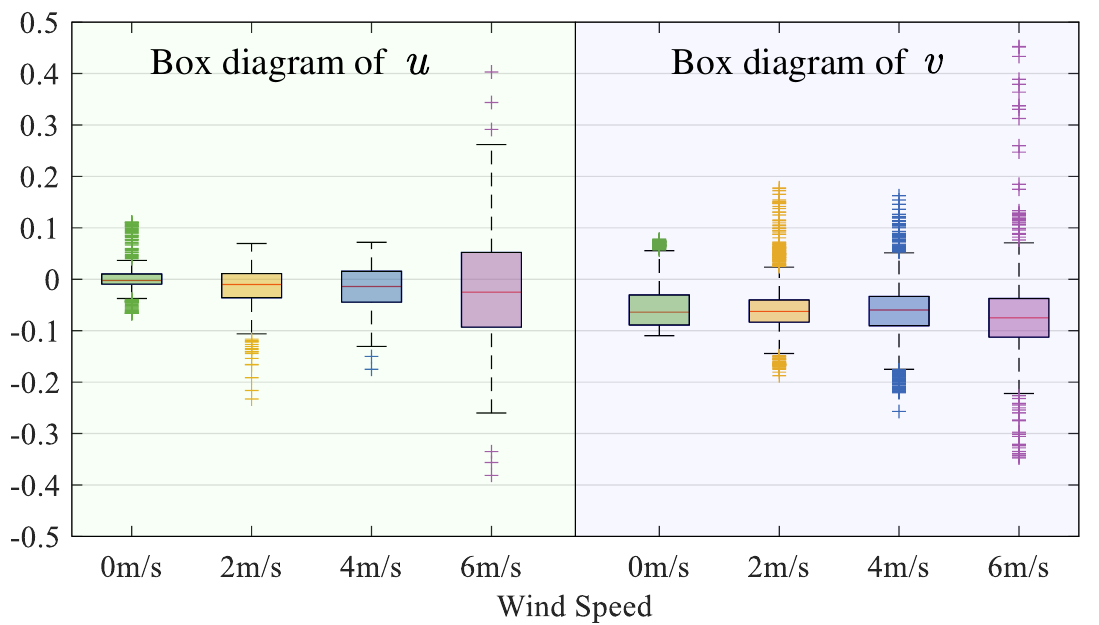}
    \caption{Comparison of normalized planar image errors at different wind speeds.}
    \label{fig:18}
\end{figure}

The median, quartiles, boundaries, and outliers of the target in the image plane are shown in Fig. \ref{fig:18}, while the Interquartile Range (IQR, defined as the difference between the upper and lower quartiles) and the interception success rate are presented in Table \ref{tab:2}. As the wind speed increases, both the median of $u$ and the IQR increase, indicating larger errors in the yaw angular rate controller. Conversely, the median and IQR of $v$ initially increase, then decrease (or vice versa), suggesting that the wind initially caused the target balloon to float upward, bringing it closer to the center of the image plane. As the wind speed continued to rise, the balloon gradually deviated from the image center. When the wind speed was below 4 m/s, the proposed controller achieved an interception success rate of over 80\%. However, at a wind speed of 6 m/s, the success rate dropped to 40\%.

Overall, although the interception success rate and robustness of the proposed system decrease as wind speed increases, it still demonstrates effectiveness in breezy conditions. Moreover, unlike balloons, real-world non-cooperative targets (e.g., drones, birds, kites, etc.) do not oscillate with the wind in the same manner.

\begin{table}[!ht]
	\centering
	\caption {Normalized image plane distribution data and interception success rate under different wind conditions.}
	\label{tab:2}
	\begin{tabular}[c]{*{4}{c}}
		\toprule
		Scenarios& median of ($u$,$v$) &IQR of ($u$,$v$)& Success rate \\
            \midrule
		{0 m/s} & (-0.002,-0.064) & (0.019,0.058) &\textbf{100\%} \\
		\midrule
	{2 m/s} & (-0.010,-0.063) & (0.046,0.043) &\textbf{80\%} \\
  
  		\midrule
		{4 m/s} & (-0.014,-0.060) &(0.061,0.057)&\textbf{80\%} \\
  		\midrule
		{6 m/s} & (-0.025,-0.075) &(0.145,0.075)&\textbf{40\%} \\
		\bottomrule
	\end{tabular}
\end{table}

\subsection{Outdoor Experiments with Different Targets}

To thoroughly evaluate the performance of the proposed controller in real-world scenarios, we designed experiments to intercept both static targets and three types of maneuvering targets in the field. The three types of maneuvering targets are designed as accelerated escape, uniform takeoff, and circular trajectory. The wind speed during the experiment ranged from 0 to 5 m/s.

\begin{figure}[!ht]
    \centering
    \includegraphics[width=0.9\linewidth]{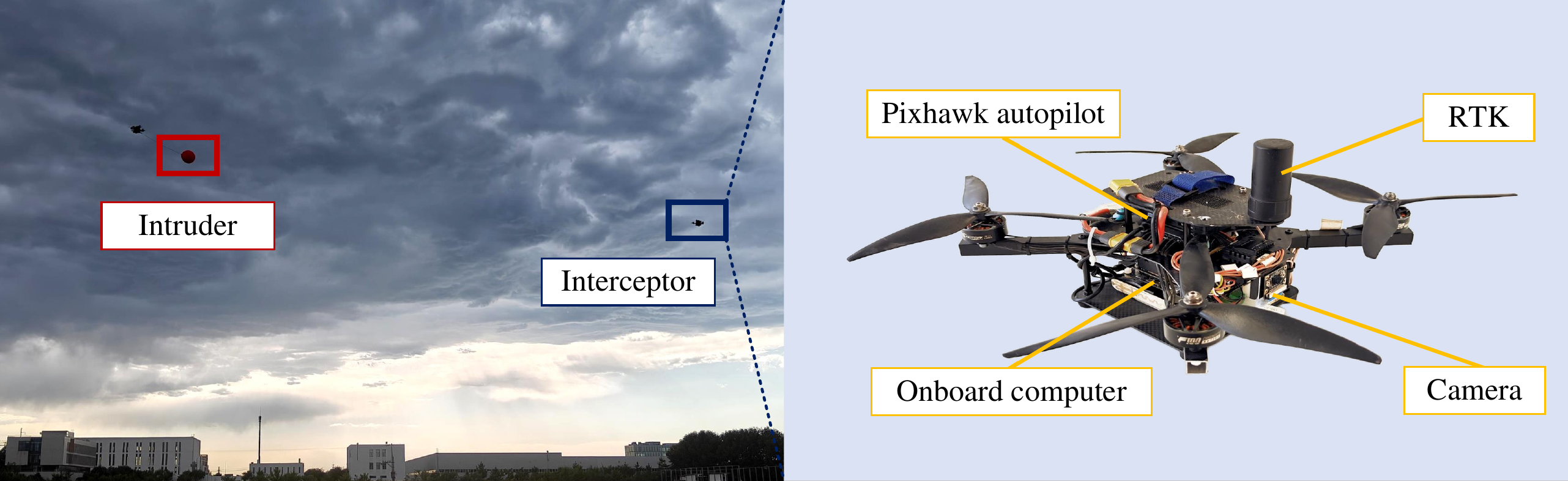}
    \caption{Real-world flight experiments and hardware architectures for intercepting drones.}
    \label{fig:10}
\end{figure}

\begin{figure}
    \centering
    \includegraphics[width=0.5\linewidth]{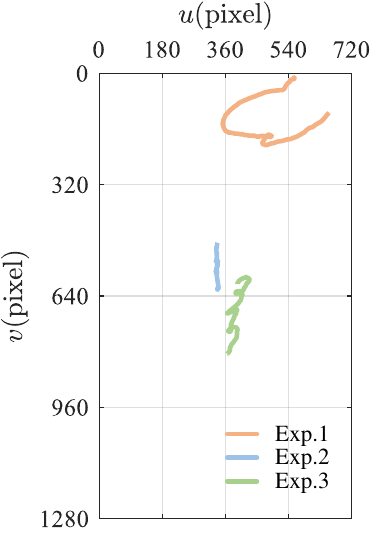}
    \caption{The trajectory of the target in image plane when intercepting a maneuvering target.}
    \label{fig:14}
\end{figure}

The outdoor experiments are shown in Fig. \ref{fig:10}. The intruder is a red balloon, similar to the one used in the indoor experiments, attached to a multicopter by a rope. The interceptor is another multicopter, equipped with various sensors and actuators. The hardware architecture of the interceptor is similar to that of the indoor experiment, featuring a larger wheelbase and a higher thrust-to-weight ratio. The multicopter has a thrust-to-weight ratio of approximately 3 and is capable of providing a maximum acceleration of 5 $\mathrm{m/s^2}$ to successfully intercept the target. Additionally, we oriented the camera horizontally, which increased the multicopter's field of view in the vertical direction, as shown in Fig. \ref{fig:14}. The interceptor multicopter start the interception task from a standstill from about 15 m away. A video of the flight experiments can be found at \href{https://youtu.be/sQdRCgnRp4g.}{https://youtu.be/sQdRCgnRp4g} or \href{https://b23.tv/nBPFA7j}{https://b23.tv/nBPFA7j}.

\begin{figure}[!ht]
    \centering
    \includegraphics[width=1\linewidth]{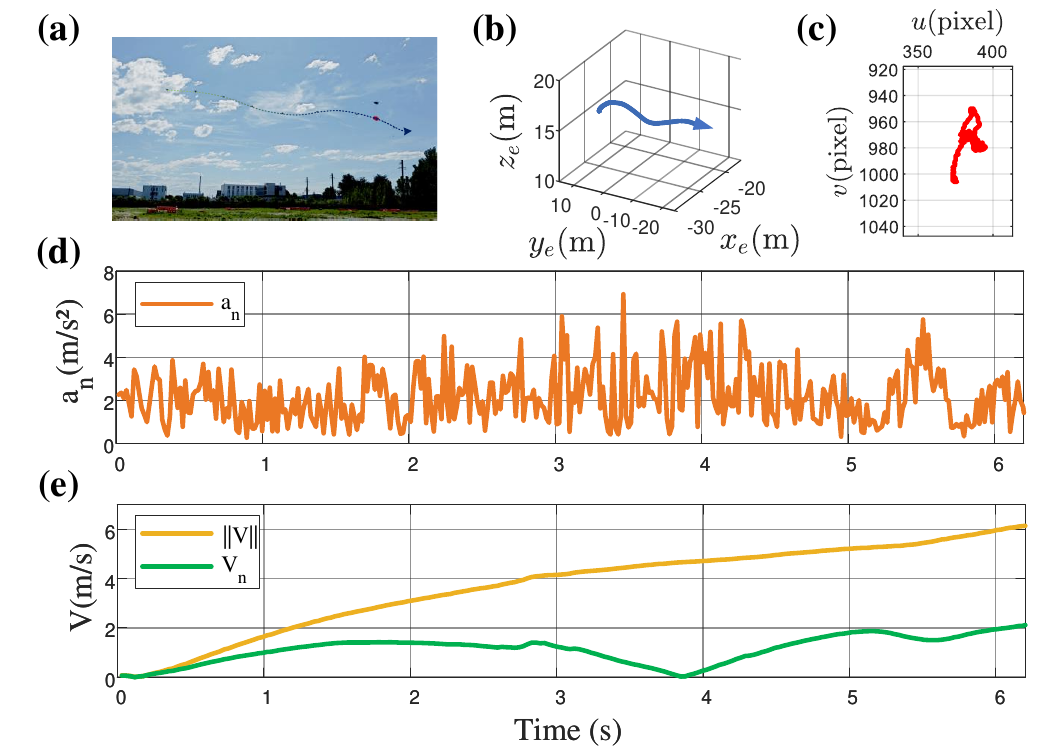}
    \caption{Results of a flight experiment to intercept a static target.}
    \label{fig:11}
\end{figure}

In the first flight experiment to intercept a static target, the interceptor successfully intercepted the target after approximately 6.2 s, reaching a terminal speed of 6.3 m/s. The results of the experiment are shown in Fig. \ref{fig:11}. Due to the field winds, the balloon experienced slight movement, requiring continuous velocity direction adjustment by the interceptor's controller, as shown in the trajectories in (a) and (b). Additionally, due to the wind, a normal acceleration of approximately 2 m/s is required to achieve hovering, as shown in (d) at $t=0$ s. (d) and (e) show that the normal acceleration increments are small and the normal velocity remains low throughout the interception process, demonstrating the effectiveness of the proposed PNG-based controller. The effectiveness of the proposed FOV holding controller is also demonstrated by the fact that the target moved only a small distance in the image plane, as shown in (c).

\begin{figure}[!ht]
    \centering
    \includegraphics[width=1\linewidth]{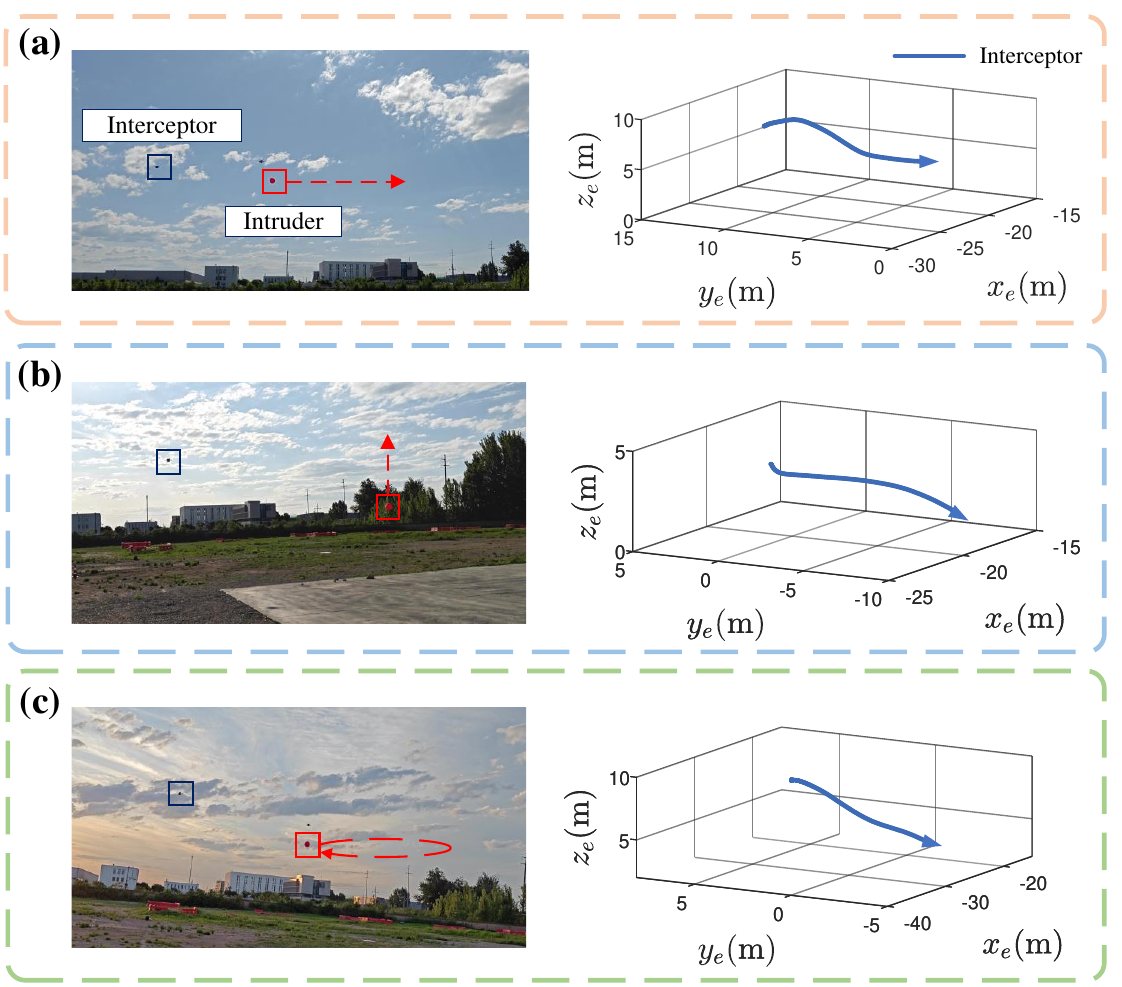}
    \caption{Schematic of (a) accelerated escape, (b) uniform takeoff, and (c) circular trajectory flight targets and interceptor trajectories.}
    \label{fig:12}
\end{figure}

\begin{figure}[!ht]
    \centering
    \includegraphics[width=1\linewidth]{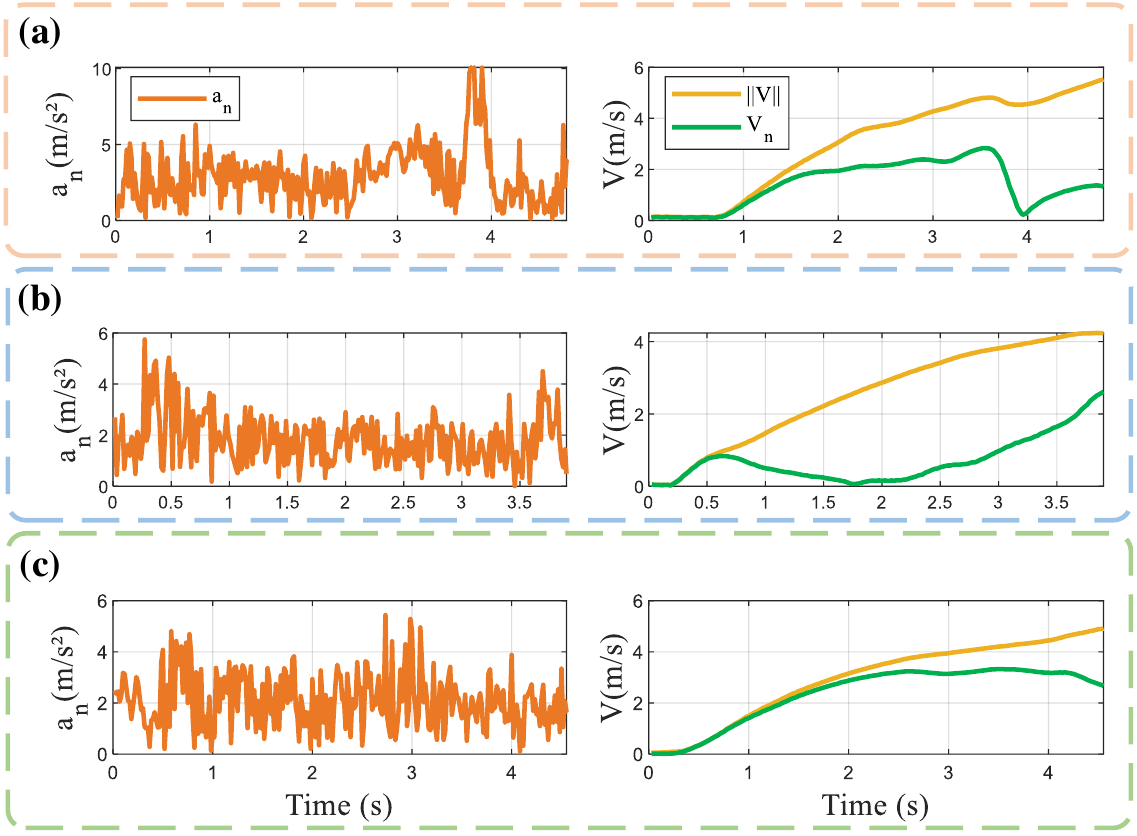}
    \caption{Normal accelerations, velocities, and normal velocities of the interceptor during the interception of (a) accelerated escape, (b) uniform takeoff, and (c) circular trajectory flight targets.}
    \label{fig:13}
\end{figure}

In the next three flight experiments intercepting maneuvering targets, the interceptor successfully intercepted the target along smooth trajectories, as shown in Fig. \ref{fig:12}. Fig. \ref{fig:13} show the velocity and normal acceleration during the interception process. During the interception of the accelerated escape target and the circular trajectory target, a significant portion of the interceptor's velocity was used for normal velocity in the early stage due to drastic changes in line-of-sight angle caused by large lateral movement of the target. In addition, the normal acceleration of interceptor at the end phase was essentially the same as that in the hovering stage. These results indicate the PNG works to ensure interception accuracy. All three scenarios achieve effective interception at a terminal velocity of about 5 m/s. Additionally, all three experiments kept the target within the field of view (FOV) throughout the interception process, as shown in Fig. \ref{fig:14}. Overall, all targets remained within a small region at the center of the image. Despite some deviation in the first experiment when intercepting the accelerated escape target, the results demonstrate the effectiveness and robustness of the proposed FOV holding controller in intercepting maneuvering targets in real scenarios.

\begin{figure*}[!ht]
    \centering
    \includegraphics[width=0.9\linewidth]{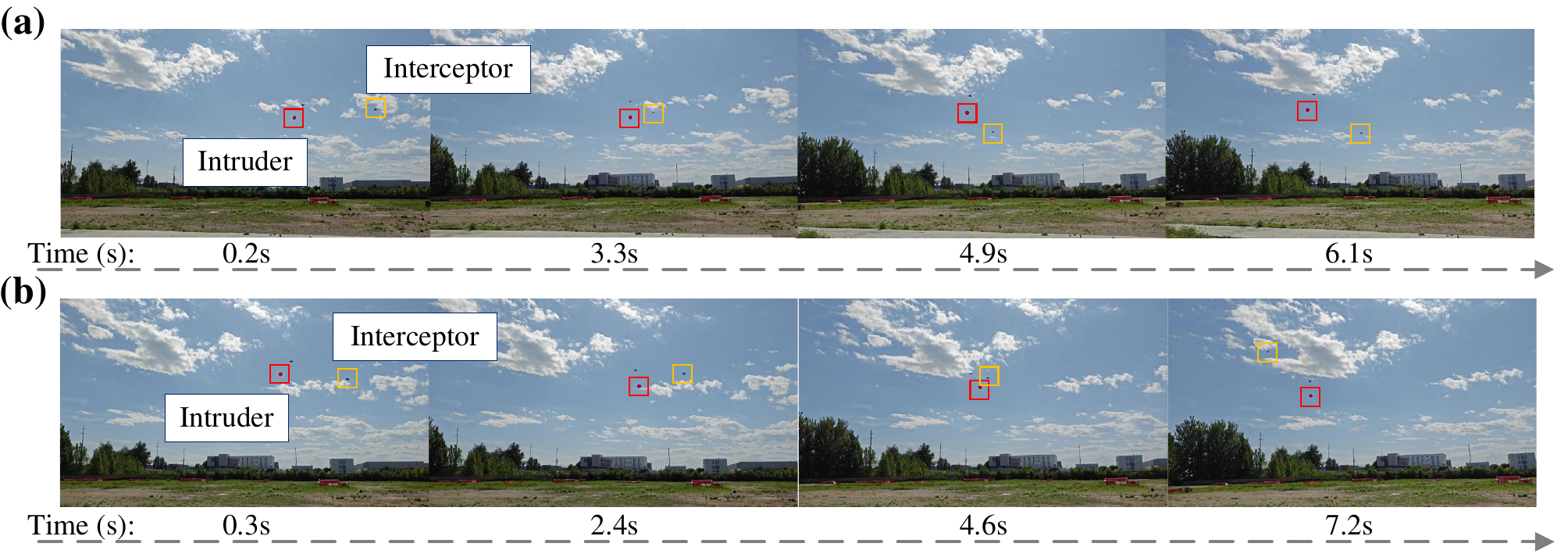}
    \caption{Snapshots of comparative challenging interception experiments for intercepting accelerated escape targets. Both contrasting algorithms of (a) \cite{paper19} and (b) \cite{paper22} failed.}
    \label{fig:15}
\end{figure*}

However, the two most relevant and comparable algorithms \cite{paper19} and \cite{paper22} were tested on intercepting an accelerated escape target. This maneuver model has been regarded as the most challenging by previous studies. As shown in Fig. \ref{fig:15}, both algorithms failed to complete the interception successfully and were manually returned by remote control after failure. Upon analysis, the reasons for the interception failure of these two algorithms are identified as: (1) The overload during interception exceeds the dynamic range of the multicopter, causing the target to move out of the FOV. (2) Random wind disturbances during the experiments cause irregular balloon oscillations. (3) The interception accuracy of the two algorithms exceeded the size of the balloon.

\section{Conclusion}
\label{sec:6}
This paper presents an advanced Image-Based Visual Servoing (IBVS) algorithm integrated with proportional navigation guidance (PNG) for the precise interception of non-cooperative aerial targets using a multicopter. The proposed algorithm leverages the flat trajectory generated by PNG, effectively mitigating errors caused by image processing latency and significantly enhancing interception accuracy. Furthermore, the incorporation of a Field of View (FOV) holding controller ensures continuous and stable target tracking, even in dynamic and unpredictable environmental conditions. Through extensive simulations and real-flight experiments under various scenarios, the proposed algorithm has demonstrated its superior performance in terms of accuracy and robustness when compared to existing methods. However, the algorithm's performance may degrade when intercepting high-speed targets due to challenges in maintaining trajectory stability at high velocities. In future work, we plan to refine the guidance law by incorporating adaptive strategies, real-time state estimation techniques, and more precise control methods to further improve the interception performance, particularly in complex and high-speed target interception scenarios.


\bibliographystyle{Bibliography/IEEEtranTIE}
\bibliography{Bibliography/IEEEabrv,Bibliography/BIB_xx-TIE-xxxx}\ 

\begin{IEEEbiography}[{\includegraphics[width=1in,height=1.25in,clip,keepaspectratio]{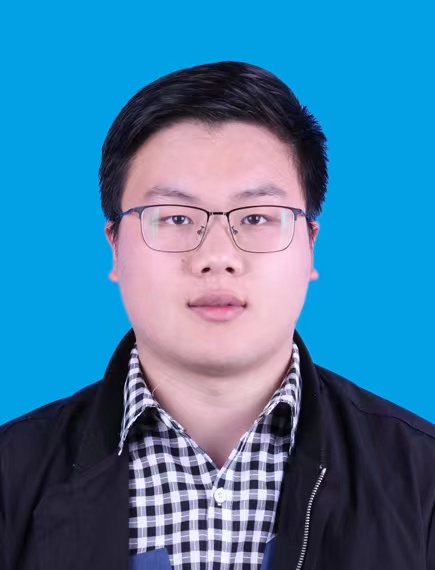}}]
{Hailong Yan} received the B.S. degree in mechanical engineering from
 University of Science and Technology Beijing, Beijing, China, in 2022.
 He is currently working toward the Ph.D.
 degree in computer science and technology with the School of Computer Science, Northwestern Polytechnical University, Xi'an, China.

 His main research interests include UAV swarm navigation, multicopter visual servo control, and cooperative guidance.
\end{IEEEbiography}

\begin{IEEEbiography}[{\includegraphics[width=1in,height=1.25in,clip,keepaspectratio]{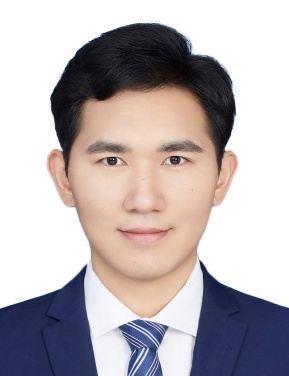}}]
{Kun Yang} is working toward the Ph.D. degree at the School of Automation Science and Electrical Engineering, Beihang University (formerly Beijing University of Aeronautics and Astronautics), Beijing, China. He received his B.S. degree in North China Electric Power University in 2018, and his M.S. degree in Beihang University in 2021.  

His main research interests include vision-based navigation and swarm confrontation.
\end{IEEEbiography}

\begin{IEEEbiography}[{\includegraphics[width=1in,height=1.25in,clip,keepaspectratio]{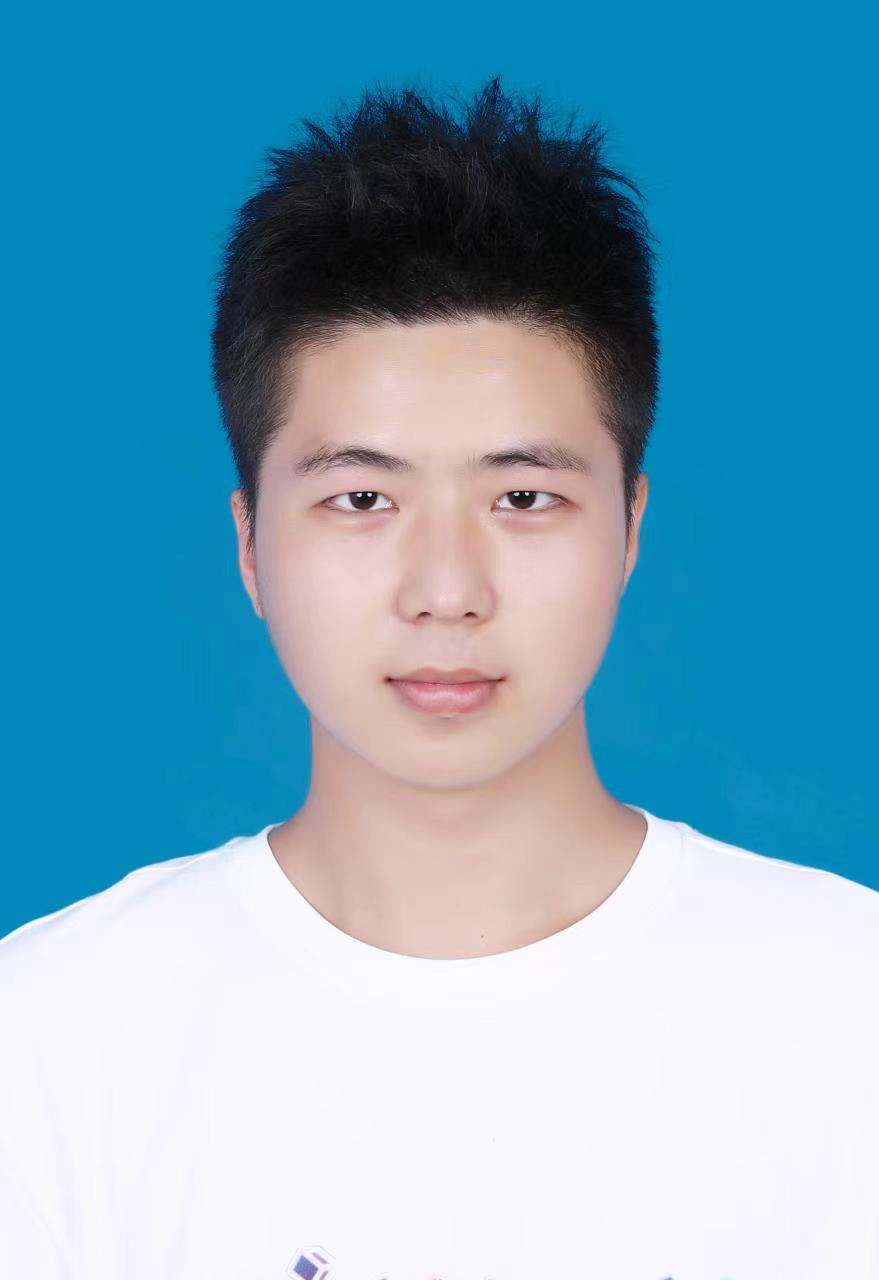}}]
{Yixiao Cheng} is currently working toward the B.S. degree in School of Future Aerospace Technology, Beihang University (formerly Beijing University of Aeronautics and Astronautics), 
 Beijing, China.

His main research interests include Synchronization Guidance Algorithm and UAV visual servo control.

\end{IEEEbiography}
\begin{IEEEbiography}[{\includegraphics[width=1in,height=1.25in,clip,keepaspectratio]{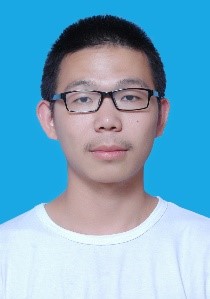}}]
{Zihao Wang} is currently pursuing a Ph.D. in Signal and Information Processing at the School of Cyber Security, University of Chinese Academy of Sciences. He received his B.S. degree in Intelligence Science and Technology from Hunan University in 2019. 

His current research interests include UAV swarm navigation, motion planning, and control.

\end{IEEEbiography}
\begin{IEEEbiography}[{\includegraphics[width=1in,height=1.25in,clip,keepaspectratio]{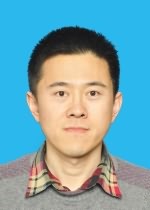}}]
{Dawei Li} received the B.S. and M.S. degrees in aeronautical and astronautical sciences and technology from Beihang University in 2001 and 2005,  respectively, and the Ph.D. degree in control science and engineering from Beihang University in 2013.

He is currently an associate professor with the Institute of Unmanned System at Beihang University. Prof. Li has a strong academic background and practical experience in the field of aircraft design and flight control. His main research interests include micro UAV design, computational fluid dynamics, flight control, and aerial swarm control and navigation.

\end{IEEEbiography}

\end{document}